\pdfoutput=1

\documentclass[11pt]{article}

\usepackage{acl}

\usepackage{times}
\usepackage{comment}
\usepackage{latexsym}
\usepackage{url}
\usepackage{graphicx}
\usepackage{amsmath}

\usepackage{subcaption,booktabs,siunitx}
\usepackage{pgfplots}
\pgfplotsset{compat=1.17} 
\usepackage[T1]{fontenc}

\usepackage[utf8]{inputenc}
\usepackage{booktabs, multirow, multicol, makecell} 
\usepackage{xcolor} 

\usepackage{microtype}
\usepackage{xspace}
\newcommand{\machamp}{{\sc MaChAmp}\xspace}
\newcommand{\xlmt}{{\sc XLM-T}\xspace}
\newcommand{\xlmr}{{\sc XLM-R}\xspace}
\definecolor{bananamania}{rgb}{0.98, 0.91, 0.71}
\expandafter\def\expandafter\UrlBreaks\expandafter{\UrlBreaks
  \do\a\do\b\do\c\do\d\do\e\do\f\do\g\do\h\do\i\do\j%
  \do\k\do\l\do\m\do\n\do\o\do\p\do\q\do\r\do\s\do\t%
  \do\u\do\v\do\w\do\x\do\y\do\z\do\A\do\B\do\C\do\D%
  \do\E\do\F\do\G\do\H\do\I\do\J\do\K\do\L\do\M\do\N%
  \do\O\do\P\do\Q\do\R\do\S\do\T\do\U\do\V\do\W\do\X%
  \do\Y\do\Z}

\usepackage{soulutf8}
\usepackage[normalem]{ulem}
\usepackage{xcolor}

\usepackage{enumitem}
\setlist{nolistsep}
\usepackage{microtype}

\iftrue 
    \usepackage[final]{proofing}
  \renewcommand\hl[1]{{#1}}  
   {\draftnote{\red{#2}}}
   \newcommand\redHL[1]{}
  \newcommand\todo[1]{}

  \newcommand{\Djame}[1]{}

\else  

\usepackage[final]{proofing}

\newcommand{\Djame}[1]{
\textbf{\textcolor{red}{\hl{Djame: #1}}}
}

\newcommand\red[1]{{{\textcolor{red}{\bf #1}}}}

\let\oldred\red
\renewcommand\red[1]{{ \oldred{{#1}}}}

 \newcommand\redHL[1]{\red{\hl{#1}}}
\let\olddraftnote\draftnote
\renewcommand\draftnote[1]{\olddraftnote{\red{#1}}}

\fi

\title{Multilingual Auxiliary Tasks Training: Bridging the Gap between Languages for Zero-Shot Transfer of Hate Speech Detection Models}

\author{Syrielle Montariol\(^{*}\) \quad Arij Riabi\(^{*}\) \quad Djamé Seddah\\
	INRIA Paris, France \\
	\texttt{firstname.lastname@inria.fr}
}

\begin{document}
\maketitle

\renewcommand\thefootnote{\fnsymbol{footnote}}
\footnotetext[1]{These authors contributed equally.}
\renewcommand*\thefootnote{\arabic{footnote}}

\begin{abstract}

Zero-shot cross-lingual transfer learning has been shown to be highly challenging for tasks involving a lot of linguistic specificities or when a cultural gap is present between languages, such as in hate speech detection. In this paper, we highlight this limitation for hate speech detection in several domains and languages using strict experimental settings. Then, we propose to train on multilingual auxiliary tasks -- sentiment analysis, named entity recognition, and tasks relying on syntactic information -- to improve zero-shot transfer of hate speech detection models across languages. We show how hate speech detection models benefit from a cross-lingual {\em knowledge proxy} brought by auxiliary tasks fine-tuning and highlight these tasks' positive impact on bridging the hate speech linguistic and cultural gap between languages.

\end{abstract}

\section{Introduction}

\iffalse
As it is clear from the daily outrages caused by racial, sexist
and political abuses on social media, offensive or hateful contents
targeted at individuals and social groups are openly circulating via online platforms. The Natural Language Processing (NLP) community has devoted considerable efforts to automatic hate speech
detection using machine learning-based approaches, and proposed
different benchmarks and datasets to evaluate them
\cite{dinakar2011modeling,Sood2012Profanity,waseem-hovy-2016-hateful,davidson2017automated,fortuna2018survey,kennedy-etal-2020-contextualizing}.
\else 

\draftadd{Given the impact social media hate speech can have on
our society as a whole -- leading to many small-scale {\em Overton window}
effects -- the NLP community has devoted considerable efforts to automatic hate speech
detection using machine learning-based approaches, and proposed
different benchmarks and datasets to evaluate their techniques
\cite{dinakar2011modeling,Sood2012Profanity,waseem-hovy-2016-hateful,davidson2017automated,fortuna2018survey,kennedy-etal-2020-contextualizing}.}
\fi

However, these systems are designed to be efficient at a given point
in time for a specific type of online content they were trained on. As hate speech varies significantly diachronically \cite{Komal2020Time} and synchronically \cite{yin2021towards}, hate speech detection models need to be constantly adapted to new contexts. 
\draftadd{For example, as noted by \citet{markov-etal-2021-exploring}, the occurrence of new hate speech domains and their associated lexicons and expressions can be triggered by real-world events, from local scope incidents to worldwide crisis.\footnote{ e.g.\ Hate speech towards  Chinese communities spiked in 2020 with the emergence of the COVID-19 Pandemic.} }
New annotated datasets are needed to optimally capture all these domain-specific, target-specific hate speech types. The possibility of creating and constantly updating exhaustively annotated datasets, adapted to every possible language and domain, is chimerical. Thus, the task of hate speech detection is often faced with low-resource issues. 

In this low-resource scenario for a given target language and domain, if annotated data is available in another language, the main option for most NLP tasks is to perform \textit{zero-shot} transfer using a multilingual language model \cite{conneau-etal-2020-unsupervised}. However, in our case, hate speech perception is highly variable across languages and cultures; for example, some slur expressions can be considered not offensive in one language, denoting an informal register nonetheless, but will be considered offensive, if not hateful, in another \cite{nozza-2021-exposing}. Despite the cross-lingual transfer paradigm being extensively used in hate speech detection to cope with the data scarcity issue \cite{basile2018crotonemilano,van-der-goot-etal-2018-bleaching,pamungkas-patti-2019-cross,ranasinghe-zampieri-2020-multilingual} \draftadd{or even the use of models trained on a translation of the initial training data \citep{rosa2021cost}} , this strong hate speech cultural and linguistic variation can lower the transferability of hate speech detection models across languages in a zero-shot setting. 

To overcome this limitation, in the absence of training data or efficient translation models for a target language, the cultural and linguistic information specific to this language needs to be found elsewhere.
In this paper, we propose to capture this information by fine-tuning the language model on resource-rich tasks in both the transfer's source and target language. Indeed, even though hate-annotated datasets are not available in both languages, it is likely that similarly annotated data in the source and target language exist for other tasks. A language model jointly fine-tuned for this other task in the two languages can learn some patterns and knowledge, bridging the gap between the languages, and helping the hate speech detection model to be transferred between them.

\draftreplace{To sum up}{In summary}, our work focuses on zero-shots cross-language multitask architectures where annotated hate speech data is available only for one source language, but some annotated data for other tasks can be accessed in both the source and target languages. Using a multitask architecture \cite{van-der-goot-etal-2021-massive} on top of a multilingual model, we investigate the impact of auxiliary tasks operating at different sentence linguistics levels (POS Tagging, Named Entity Recognition (NER), Dependency Parsing and Sentiment analysis) on the transfer effectiveness. Using
\citet{nozza-2021-exposing}'s original set of languages and datasets
(hate speech against women and immigrants, from Twitter datasets in English, Italian and Spanish), our main contributions are as follows.
\begin{itemize}
\item Building strictly comparable corpora across languages,\footnote{Our comparable datasets are available at \url{https://github.com/ArijRB/Multilingual-Auxiliary-Tasks-Training-Bridging-the-Gap-between-Languages-for-Zero-Shot-Transfer-of-/}.}  leading to a thorough evaluation framework, we highlight cases where zero-shot cross-lingual transfer of hate speech detection models fails and diagnose the effect of the choice of the multilingual language model.
\item We identify auxiliary tasks with a positive impact on cross-lingual transfer when trained jointly with hate speech detection: sentiment analysis and NER. The impact of syntactic tasks is more mitigated.
\item Using the HateCheck test suite \cite{rottger-etal-2021-hatecheck, rottger-etal-2022-multilingual}, we identify which hate speech \textit{classes of functionalities} suffer the most from cross-lingual transfer, highlighting the impact of \textit{slurs}; and which ones benefit from joint training with multilingual auxiliary tasks.
\end{itemize}

\section{Related Work}

\paragraph{Intermediate task training.}

In order to improve the efficiency of a pre-trained language model for a given task, this model can undergo preliminary fine-tuning on an intermediate task before fine-tuning again on the downstream task. This idea was formalized as Supplementary Training on Intermediate Labeled-data Tasks (STILT) by \citet{phang2018sentence}, who perform sequential task-to-task pre-training. 
More recently, \citet{pruksachatkun-etal-2020-intermediate} perform a survey of intermediate and target task pairs to analyze the usefulness of this intermediary fine-tuning, but only in a monolingual setting. 
\citet{phang-etal-2020-english} turn towards cross-lingual STILT. They fine-tune a language model on nine intermediate language-understanding tasks in English and apply it to a set of non-English target tasks. They show that machine-translating intermediate task data for training or using a multilingual language model does not improve the transfer compared to English training data. However, to the best of our knowledge, using intermediate task training data on both the source and the target language for transfer has not been tested in the literature.

\paragraph{Auxiliary tasks for hate speech detection.}

Auxiliary task training for hate speech detection has been done almost exclusively with the sentiment analysis task \cite{Bauwelinck2019Measuring, aroyehun2021evaluation}, and only in monolingual scenarios. But additional information is sometimes added to the hate speech classifier differently.
\citet{gambino2020chilab}, among the best systems on the HaSpeeDe task of EVALITA 2020, use POS-tagged text as input of the classification systems, which is highly beneficial for Spanish and a bit less for German and English. Furthermore, the effect of syntactic information is also investigated by \citet{narang-brew-2020-abusive}, using classifiers based on the syntactic structure of the text for abusive language detection. \citet{markov-etal-2021-exploring} evaluate the impact of manually extracted POS, stylometric and emotion-based features on hate speech detection, showing that the latter two are robust features for hate speech detection across languages.

\paragraph{Zero-shot cross-lingual transfer for hate speech detection}
Due to the lack of annotated data on many languages and domains for hate speech detection, zero-shot cross-lingual transfer has been tackled a lot in the literature. Among the most recent work, \citet{pelicon2021investigating} investigates the impact of a preliminary training of a classification model on hate speech data languages different from the target language; they show that language models pre-trained on a small number of languages benefit more of this intermediate training, and often outperforms massively multilingual language models. 
To perform cross-lingual experiment, \citet{glavas-etal-2020-xhate} create a dataset with aligned examples in six different languages, avoiding the issue of hate speech variation across languages that we tackle in this paper. On their aligned test set, they show the positive impact of intermediate masked language model fine-tuning on abusive corpora in the target language. Using aligned corpora allows the authors to focus on the effect of the intermediate finetuning without the noise of inter-language variability. On the contrary, in our case, we investigate the issue of limited transferability of hate speech detection models across languages. 
\citet{nozza-2021-exposing}, on which this paper builds upon, demonstrates the limitation of cross-lingual transfer for domain-specific hate speech -- in particular, hate speech towards women -- and explains it by showing examples of cultural variation between languages. Some notable hate speech vocabulary in one language may be used as an intensifier in another language.\footnote{\citet{nozza-2021-exposing} gives the example of the Spanish word \textit{puta} often used as an intensifier without any misogynistic connotation, while it translates to a slang version of ``prostitute'' in English.} \citet{stappen2020cross} perform zero- and few-shots cross-lingual transfer on some of the datasets we use in this paper, with an attention-based classification model; but contrarily to us, they do not distinguish \draftadd{between}  the hate speech targets.

\section{The Bottleneck of Zero-shot Cross-lingual Transfer}\label{sec:bottleneck}

\subsection{Hate speech corpora}

We use the same hate speech datasets as \citet{nozza-2021-exposing}, who relied on them to point out the limitations of zero-shot cross-lingual transfer. The corpora are in three languages: English (en), Spanish (es) and Italian (it); and two domains: hate speech towards immigrants and hate speech towards women. 
The corpora come from various shared tasks; For English and Spanish, we use the dataset from a shared task on hate speech against immigrants and
women on Twitter (HatEval). For the Italian corpora, we use the automatic misogyny identification challenge (AMI) \cite{fersini2018overview} for the women domain and the hate speech detection shared task on Facebook and Twitter (HaSpeeDe) \cite{bosco2018overview} for the immigrants domain. Links to the resources are listed in Table \ref{tab:shared_hate} in Appendix \ref{sec:appendix-datasets}.

The hate speech detection task is a binary classification task where each dataset is annotated with two labels: \textit{hateful} and \textit{non hateful}. 
We train binary classification models on the train sets in each language and predict on the test set of each language, investigating two settings: 1) monolingual, i.e, training and testing on the same language and domain for hate speech; 2) zero-shot, cross-lingual, i.e.\ training on one and testing on another. We evaluate the models using macro-F1 as metric.

\subsection{Original baseline results}

The original results reported by \citet{nozza-2021-exposing} can be found in the first rows of Table \ref{tab:baseline_reslts_main}. In the table, we highlight in brown zero-shot cross-lingual cases where the macro-F1 score drops by more than 25\% compared to the monolingual setting: these are cases for which we consider that the cross-lingual transfer failed.
We observe the phenomenon that raised the issue of zero-shot cross-lingual transfer: in the \textit{women} domain, the models trained on Spanish and Italian in a zero-shot setting have much lower scores compared to the monolingual results; 4 out of the 6 cross-lingual cells are highlighted in brown. One possible cause, as explained by \citet{nozza-2021-exposing}, is the presence of language-specific offensive interjections that lead the model to wrongly classify text as hateful towards women. 

On a side note, models trained and tested on the English corpus on the immigrants domain have particularly low scores (macro-F1 of 36.8 in the monolingual setting). This phenomenon was also observed by \citet{nozza-2021-exposing} and \citet{stappen2020cross}, and is explained by the authors by the presence of specific words and hashtags that were used for scraping the tweets and that lead the model to over-fit, linked with a large discrepancy between the train and test set.

\begin{table}
\footnotesize
\begin{tabular}{@{}p{1.1cm}p{0.4cm}cccccc@{}}\toprule
\multirow{2}{*}{\centering Model} & \multirow{2}{*}{\parbox{0.5cm}{Src lang}} & \multicolumn{3}{c}{immigrants} & \multicolumn{3}{c}{women} \\
\cmidrule(lr){3-5}
\cmidrule(l){6-8}
 & &en &es &it &en &es &it \\\midrule
 \multirow{3}{*}{\makecell{m-BERT\\ \scriptsize{\citet{nozza-2021-exposing}}}}
  &en &36.8 &63.3 &59.0 &55.9 & 54.6 & 44.9 \\
 &es & 59.6 & 63.0 &68.3 &\textcolor{orange!80!black}{55.8} &83.9 &\textcolor{orange!80!black}{33.7} \\
 &it &63.5 & 66.6 &77.7 & \textcolor{orange!80!black}{54.5} &\textcolor{orange!80!black}{46.3} &80.8 \\
\midrule
\multicolumn{8}{c}{Comparable corpus size and new random split}\\
\midrule
\multirow{3}{*}{m-BERT }& en  & 72.5 & \textcolor{orange!80!black}{48.5} & 63.8 & 75.2 & \textcolor{orange!80!black}{41.7} & \textcolor{orange!80!black}{43.4} \\ 
 & es  & \textcolor{orange!80!black}{59.4} & 80.9 & \textcolor{orange!80!black}{58.5} & \textcolor{orange!80!black}{54.5} & 76.9 & \textcolor{orange!80!black}{40.5} \\ 
 & it  & 62.8 & \textcolor{orange!80!black}{54.8} & 76.3 & \textcolor{orange!80!black}{46.3} & \textcolor{orange!80!black}{53.6} & 88.3 \\ 
 \midrule
 \multirow{3}{*}{\xlmr }& en  & 75.3 & \textcolor{orange!80!black}{51.9} & 70.1 & 76.6 & \textcolor{orange!80!black}{51.6} & \textcolor{orange!80!black}{49.9} \\ 
 & es  & \textcolor{orange!80!black}{62.0} & 83.4 & 65.4 & 63.4 & 77.8 & \textcolor{orange!80!black}{46.9} \\ 
 & it  & 69.2 & \textcolor{orange!80!black}{51.3} & 78.6 & \textcolor{orange!80!black}{60.3} & \textcolor{orange!80!black}{57.3} & 89.0 \\ 
 \midrule
  \multirow{3}{*}{\xlmt} & en  & 76.8 & \textcolor{orange!80!black}{48.5} & 73.5 & 78.6 & 61.5 & 60.6 \\ 
 & es  & 65.9 & 84.2 & \textcolor{orange!80!black}{60.7} & 72.5 & 80.3 & \textcolor{orange!80!black}{51.9} \\ 
 & it  & 71.5 & \textcolor{orange!80!black}{56.8} & 78.4 & \textcolor{orange!80!black}{63.4} & \textcolor{orange!80!black}{58.2} & 90.3 \\ 
\bottomrule
\end{tabular}


\caption{Monolingual and cross-lingual hate speech detection macro-F1 scores on all corpora. All results except for the one from \citet{nozza-2021-exposing} are macro-F1 (\%) averaged over 5 runs. All use 20 epochs. Numbers in \textcolor{orange!80!black}{brown} highlight cases when the loss in performance in the zero-shot cross-lingual case compared to the monolingual case is higher than 25\%.}
\label{tab:baseline_reslts_main}
\end{table}

\subsection{Experimental settings}\label{sec:experimental-settings}

\paragraph{Building comparable corpora.} 
We started this work to investigate the failure of cross-lingual hate speech datasets for the women domain highlighted by \citet{nozza-2021-exposing}. However, these experiments were not realized in comparable settings; the corpora do not have the same size in the different languages and domains. Our goal is to confirm these results under a strictly comparable setting, and a multi-seed robust experimental framework. Therefore, we build comparable corpora in each language and domain to ensure the comparability of the transfer settings. We reduce all datasets to a total size of 2\,591 tweets, the size of the smallest one, sampling from each original split separately; each train set has 1\,618 tweets, each development set 173, and each test set 800.
We use the Kolmogorov--Smirnov test to compare the sentence length distribution (number of tokens) and the percentage of hate speech between the sampled and the original datasets, to make sure they stay comparable. The sampling is done randomly until the similarity conditions with the original dataset are met.
The original size for each dataset as well as the sampling size for building the comparable datasets and the percentage of hateful examples can be found in Table \ref{tab:hatespeech_data_size} and Table \ref{tab:hatespeech_percentage} in Appendix \ref{sec:appendix-datasets}.

On top of this, before the sub-sampling of the corpora, we merge the development, test and train dataset for each language and domain before performing a new random split. This allows us to overcome the train-test discrepancy observed in the English-immigrants dataset \draftadd{we mentioned above}.

\paragraph{Pre-processing.}
We process the datasets by replacing all mentions and URLs with specific tokens, and segmenting the hashtags into words.\footnote{Using the Python package \texttt{wordsegment}.} 
Given the compositional nature of hashtags (a set of concatenated words), hashtag segmentation is frequently done as a pre-processing step in the literature when handling tweets (e.g. \cite{rottger-etal-2021-hatecheck}); it can improve tasks such as tweet clustering \citep{gromann2017hashtag}.

\paragraph{Models training.}

For all our experiments, we use the \machamp v0.2 framework\footnote{\url{https://github.com/machamp-nlp/machamp}, under the MIT license.} \cite{van-der-goot-etal-2021-massive}, a multi-task toolkit based on AllenNLP \cite{gardner-etal-2018-allennlp}. 
We keep most of the default hyperparameters of \machamp for all experiments, which the authors optimized on a wide variety of tasks. 
We fine-tune a multilingual language model on the hate speech detection task for each of the six training corpora described in the previous section. We keep the best out of 20 epochs for each run according to the macro-F1 score on the development set.

Note that the new comparable test sets sampled from the original corpora are relatively small (800 observations). To increase the robustness of the results, we use five different seeds when fine-tuning a language model on the hate speech detection task and report the average macro-F1 over the five runs. 

\paragraph{Language Models.}

We use two general-domain large-scale multilingual language models: m-BERT \cite{devlin-etal-2019-bert} following \citet{nozza-2021-exposing} and \xlmr \cite{conneau-etal-2020-unsupervised}. The former is the multilingual version of BERT, trained on Wikipedia content in 104 languages, with 100M parameters. The latter has the same architecture as RoBERTa \citep{liu2019roberta} with 550M parameters and is trained on the publicly available 2.5\,TB CommonCrawl Corpus, covering 100 languages.

Then, we experiment with \xlmt \cite{barbieri2021xlmtwitter}, an off-the-shelf \xlmr model fine-tuned on 200 million tweets (1\,724 million tokens) scraped between 05/2018 and 03/2020, in more than 30 languages, including our three target languages.

\subsection{Setting a new baseline}

We compare the scores for m-BERT from \citet{nozza-2021-exposing} to the scores obtained using our comparable corpora, reported in Table \ref{tab:baseline_reslts_main}. 
First, our experiment with m-BERT on comparable corpora allows us to highlight additional cases where zero-shot cross-lingual transfer ``fails'' (macro-F1 dropping by more than 25\% compared to monolingual score) in the \textit{immigrants} domain, that were not visible in the previous study due to variations in training corpus size.
On top of this, with the new splits, we do not observe the extremely low scores on English for the immigrant domain anymore, allowing us to draw more reliable conclusions on the monolingual/cross-lingual performance gap.

Comparing m-BERT and \xlmr, the latter shows higher scores for almost all languages and domains. It also shows, in general, slightly lower macro-F1 loss between monolingual and cross-lingual settings; which is related to its much larger number of parameters and training corpus size compared to m-BERT.

Fine-tuning \xlmt leads to higher macro-F1 scores for almost all languages and domains compared to \xlmr; which is expected, as it was fine-tuned using the Masked Language Modeling (MLM) task on tweets, which is much more similar to the hate speech datasets, at least stylistically due to the Twitter platform constraints (e.g. number of characters). 
In terms of monolingual/cross-lingual discrepancy, we also observe in general a much lower macro-F1 drop. Having seen a large amount of similar data in all languages, the model can much more easily bridge the gap between languages when performing zero-shot cross-lingual transfer for this highly domain-specific task.

However, such a large amount of training data from a similar source in different languages is not so easy to come by. To bridge the language gap in very context-specific tasks such as hate speech detection, in the case of absence of an adequately trained multilingual language model, we turn towards other sources of multilingual information for the model: using annotated corpora for other \textit{auxiliary} tasks in the source and target languages.

In all following experiments, we use the comparable datasets and the general-domain multilingual language model \xlmr to study the impact of auxiliary task training on this problem\footnote{The results for \xlmt display similar tendencies with higher scores compared to \xlmr, Detailed and summarized tables can be found in 499
Appendix \ref{app:results}, Table \ref{tab:xlmt-all}}. By using data for auxiliary tasks in both the source and the target language, we expect the auxiliary task training to work as a bridge between the source and target language, helping the cross-lingual transfer by providing more information on the target language and the difference between the two languages.


\section{Auxiliary Tasks Experiments}\label{sec:aux-tasks}

We define several training tasks whose effects on cross-lingual transfer of hate speech detection models are to be evaluated: a sequence-level task, sentiment analysis, and several token-level tasks: Named Entity Recognition (NER) and a set syntactic tasks that we \draftreplace{gather}{group} -- by misnomer -- under the term ``Universal Dependency'' (UD). We hypothesize that sentiment analysis and NER tasks allow the model to learn high-level, semantic information, while the UD tasks convey syntactic skills to the model.

\subsection{Auxiliary tasks}

\paragraph{Syntactic tasks.} 

We investigate the effect of adding syntactic information by using all Universal Dependency (UD, \citealp{nivre-etal-2020-universal}) tasks (Dependency Parsing,  Part-Of-Speech (POS) tagging, lemmatization and morphological tagging).
We use the dataset EWT \cite{silveira-etal-2014-gold}, GSD and ISDT \cite{bosco2014evalita}, for English, Spanish and Italian respectively.
The datasets being of different sizes, we sample them to obtain the same training size in all languages. We use a train set size of 12\,543 sentences, the size of the smallest dataset. Detailed statistics about the datasets can be found in Table \ref{tab:complementary-datasets-ud} in Appendix \ref{sec:appendix-datasets}.

\paragraph{Sentiment analysis.}

We use Twitter sentiment analysis datasets on each of our three target languages. They have been gathered and unified by \citet{barbieri2021xlmtwitter}, with a unique split size (training 1\,839, development 324, test 870) and a balanced distribution across the three sentiment labels (positive, negative and neutral)\footnote{\url{https://github.com/cardiffnlp/xlm-t}}. Detailed statistics and additional information on each dataset can be found in Table \ref{tab:complementary-datasets-sentiment} in Appendix \ref{sec:appendix-datasets}.

\paragraph{Named Entity Recognition (NER).}
An advantage of this task, which consists in identifying entities in a sequence, is that it is more language-agnostic than the others. Indeed, named entities are often transparent between languages, making it a good choice for cross-lingual transfer.
We use the NER WikiANN dataset from \cite{pan-etal-2017-cross, rahimi-etal-2019-massively}, which covers our three languages. The sets have a unique split size (training 20k examples, development 10k, test 10k).

\subsection{Multi-task learning pipeline}

We perform multi-task learning using the \machamp framework \cite{van-der-goot-etal-2021-massive}; it fine-tunes contextual embeddings for several tasks and several datasets using a shared encoder and different decoders depending on the target task. As the datasets associated with the different tasks have varying sizes, we use a ``smooth sampling'' method to avoid having under-represented datasets during training. It consists of re-sampling the datasets according to a multinomial distribution for each batch.

We fine-tune the multilingual model \xlmr on the different auxiliary tasks. The training is done jointly on the auxiliary task datasets in the three languages, in order to allow the model to learn patterns between languages, and on the hate speech dataset in the \textit{source} language, before being tested on the \textit{target} language. In practice, the language model can be trained on the auxiliary tasks either in an intermediary fashion before being fine-tuned on the downstream task (similarly to \citet{pruksachatkun-etal-2020-intermediate}), or jointly with the hate speech detection task. According to our experiments, the latter exhibits the best performance; we report only results with joint training in the paper. All results involving hate speech are obtained using the pipeline described in Section \ref{sec:experimental-settings}, averaging the macro-F1 over five different runs.

\section{Results on Auxiliary Tasks Training}\label{sec:results}

\begin{table}[!ht]
\footnotesize
\begin{subtable}{1\linewidth}
\sisetup{table-format=-1.2}   
\centering
   
\begin{tabular}{@{}p{0.4cm}p{0.4cm}c@{\hspace{3mm}}c@{\hspace{3mm}}cc@{\hspace{3mm}}c@{\hspace{3mm}}c@{}}
    \toprule

\multirow{2}{*}{\parbox{0.5cm}{\centering Aux. task}} & \multirow{2}{*}{\parbox{0.5cm}{\centering  Src lang}} &\multicolumn{3}{c}{immigrants} & \multicolumn{3}{c}{women} \\ \cmidrule(lr){3-5}
\cmidrule(l){6-8}
& &en &es &it &en &es &it \\ \midrule
 \multirow{3}{*}{None }& en  & \textit{75.3} & \textcolor{orange!80!black}{51.9} & 70.1 & \textit{76.6} & \textcolor{orange!80!black}{51.6} & \textcolor{orange!80!black}{49.9} \\ 
 & es  & \textcolor{orange!80!black}{62.0} & \textit{83.4} & 65.4 & 63.4 & \textit{77.8} & \textcolor{orange!80!black}{46.9} \\ 
 & it  & 69.2 & \textcolor{orange!80!black}{51.3} & \textit{78.6} & \textcolor{orange!80!black}{60.3} & \textcolor{orange!80!black}{57.3} & \textit{89.0} \\ \midrule
\multirow{3}{*}{\makecell{Sent-\\ iment}}  & en  & \textcolor{red!60!black}{-1.0\hphantom{$^\dagger$}} & \textcolor{red!60!black}{-1.2\hphantom{$^\dagger$}} & \textcolor{black!60!black}{0.0\hphantom{$^\dagger$}} & \textcolor{green!60!black}{2.0$^\dagger$} & \textcolor{green!60!black}{0.9\hphantom{$^\dagger$}} & \textcolor{red!60!black}{-6.2$^\dagger$} \\ 
 & es  & \textcolor{green!60!black}{5.1$^\dagger$} & \textcolor{green!60!black}{0.6\hphantom{$^\dagger$}} & \textcolor{green!60!black}{1.5\hphantom{$^\dagger$}} & \textcolor{green!60!black}{0.7\hphantom{$^\dagger$}} & \textcolor{green!60!black}{2.1$^\ddagger$} & \textcolor{red!60!black}{-9.6$^\ddagger$} \\ 
 & it  & \textcolor{green!60!black}{1.4$^\dagger$} & \textcolor{green!60!black}{1.7\hphantom{$^\dagger$}} & \textcolor{red!60!black}{-0.9\hphantom{$^\dagger$}} & \textcolor{red!60!black}{-8.3$^\ddagger$} & \textcolor{red!60!black}{-0.7\hphantom{$^\dagger$}} & \textcolor{green!60!black}{0.1\hphantom{$^\dagger$}} \\ 

\midrule 
\multirow{3}{*}{NER} & en  & \textcolor{green!60!black}{1.4$^\dagger$} & \textcolor{green!60!black}{1.0\hphantom{$^\dagger$}} & \textcolor{red!60!black}{-1.9\hphantom{$^\dagger$}} & \textcolor{green!60!black}{0.4\hphantom{$^\dagger$}} & \textcolor{green!60!black}{0.2\hphantom{$^\dagger$}} & \textcolor{green!60!black}{1.9\hphantom{$^\dagger$}} \\ 
 & es  & \textcolor{green!60!black}{3.1\hphantom{$^\dagger$}} & \textcolor{green!60!black}{0.4\hphantom{$^\dagger$}} & \textcolor{red!60!black}{-1.1\hphantom{$^\dagger$}} & \textcolor{red!60!black}{-8.7$^\dagger$} & \textcolor{green!60!black}{2.2$^\ddagger$} & \textcolor{red!60!black}{-4.9\hphantom{$^\dagger$}} \\ 
 & it  & \textcolor{green!60!black}{3.3$^\ddagger$} & \textcolor{green!60!black}{4.5$^\ddagger$} & \textcolor{red!60!black}{-1.4$^\dagger$} & \textcolor{red!60!black}{-2.8$^\dagger$} & \textcolor{red!60!black}{-0.5\hphantom{$^\dagger$}} & \textcolor{green!60!black}{1.1$^\dagger$} \\ 

\midrule 
\multirow{3}{*}{UD} & en  & \textcolor{green!60!black}{1.7$^\dagger$} & \textcolor{red!60!black}{-2.4\hphantom{$^\dagger$}} & \textcolor{red!60!black}{-1.2\hphantom{$^\dagger$}} & \textcolor{green!60!black}{0.7\hphantom{$^\dagger$}} & \textcolor{red!60!black}{-0.4\hphantom{$^\dagger$}} & \textcolor{red!60!black}{-10.6$^\dagger$} \\ 
 & es  & \textcolor{red!60!black}{-3.6\hphantom{$^\dagger$}} & \textcolor{red!60!black}{-1.1\hphantom{$^\dagger$}} & \textcolor{red!60!black}{-6.5$^\dagger$} & \textcolor{red!60!black}{-4.9\hphantom{$^\dagger$}} & \textcolor{red!60!black}{-0.4\hphantom{$^\dagger$}} & \textcolor{red!60!black}{-10.9$^\ddagger$} \\ 
 & it  & \textcolor{red!60!black}{-14.4$^\ddagger$} & \textcolor{green!60!black}{5.0$^\ddagger$} & \textcolor{red!60!black}{-1.6$^\dagger$} & \textcolor{red!60!black}{-14.7$^\ddagger$} & \textcolor{red!60!black}{-5.6\hphantom{$^\dagger$}} & \textcolor{red!60!black}{-0.3\hphantom{$^\dagger$}} \\

\bottomrule
    \end{tabular}

   \caption{Detailed view.}\label{tab:xlmr-detail}
\end{subtable}

\bigskip
\begin{subtable}{1\linewidth}
\sisetup{table-format=4.0} 
\centering
    \begin{tabular}{lrrrr}
    \toprule 
    \multirow{2}{*}{\makecell{Auxiliary\\ Task}} &  \multicolumn{2}{c}{immigrants} &  \multicolumn{2}{c}{women} \\ 
    \cmidrule(lr){2-3}
    \cmidrule(lr){4-5}
     & mono & cross & mono & cross \\  
      \midrule 
 None  & 79.1 & 61.6 & 81.1 & 54.9 \\ 
Sentiment  & \textcolor{red!60!black}{-0.4} & \textcolor{green!60!black}{1.4} & \textcolor{green!60!black}{1.4} & \textcolor{red!60!black}{-3.9} \\ 
NER  & \textcolor{green!60!black}{0.1} & \textcolor{green!60!black}{1.5} & \textcolor{green!60!black}{1.3} & \textcolor{red!60!black}{-2.5} \\ 
UD  & \textcolor{red!60!black}{-0.3} & \textcolor{red!60!black}{-3.8} & \textcolor{black!60!black}{0.0} & \textcolor{red!60!black}{-7.8} \\ 
Sentiment + NER & \textcolor{green!60!black}{0.4} & \textcolor{green!60!black}{2.5} & \textcolor{green!60!black}{1.3} & \textcolor{red!60!black}{-4.7} \\ 
\bottomrule 

    \end{tabular}%
\caption{Aggregated view.}
\label{tab:xlmr-summary}
\end{subtable}

    \caption[]{Effect (delta with hate speech detection baseline, averaged over 5 runs) of fine-tuning \xlmr on the three auxiliary tasks, on hate speech detection macro-F1 scores (\%). \textcolor{green!60!black}{Green} values indicate an increase in score, \textcolor{red!60!black}{red} values a decrease. The subscript indicates whether the score is significantly higher or lower compared to the baseline. The comparison is made using a one-sided $t$-test over the list of scores of the five runs of each model.\protect\footnotemark A dagger ($\dagger$) as exponent indicates that the $p$-value is smaller than 0.05, while a double-dagger ($\ddagger$) indicates a $p$-value smaller than 0.01.}
    \label{tab:xlmr-all}
\end{table}

\footnotetext{\url{https://docs.scipy.org/doc/scipy/reference/generated/scipy.stats.ttest_ind.html}.}

We analyze the training effect of adding different auxiliary tasks on top of \xlmr, jointly with monolingual hate speech detection. Results can be found in Table \ref{tab:xlmr-all}. Instead of raw scores, we compute the deltas between the baseline system (no auxiliary task, same as Table \ref{tab:baseline_reslts_main}) and the augmented system with training jointly with auxiliary tasks: NER, sentiment analysis (\textit{Sent}) and syntactic tasks (UD), for each language pair (Table \ref{tab:xlmr-detail}).

To help with the interpretation, we aggregate the results according to the monolingual (\textit{mono}), and zero-shot cross-lingual (\textit{cross}) settings. Table \ref{tab:xlmr-summary} is the aggregated equivalent of Table \ref{tab:xlmr-detail}. For each domain (immigrants and women), we average the scores by setting: the \textit{mono} columns show the average of all scores in the diagonal in Table \ref{tab:xlmr-detail}, while the \textit{cross} column is the average of all the rest.

In the zero-shot cross-lingual transfer scenario, we hypothesized that the additional information on the source and target languages could bridge the gap between the languages and improve the transfer for hate speech detection. Looking at the scores for cross-lingual transfer, sentiment analysis and NER lead to an average improvement of respectively of 1.42 and 1.48 \draftadd{points} for the immigrants domains; combined (last row of Table \ref{tab:xlmr-summary}), they lead to an even greater improvement of 2.5\draftreplace{\%}{ percentage points}. On the contrary, for the women domain, these two tasks lead to significant improvements almost only in the monolingual setting. As underlined before, zero-shot cross-lingual transfer is especially hard in this domain due to cultural and linguistic variations \cite{nozza-2021-exposing} that auxiliary task training fails to capture. 
Finally, UD tasks auxiliary training leads to a large drop of performance in most cases.
The impact of auxiliary tasks on the performance of hate speech detection using the \xlmt model is comparable to the one observed with \xlmr. Detailed and summarized tables can be found in Appendix \ref{app:results}, Table \ref{tab:xlmt-all}.

\section{Diagnosis: Effect of Auxiliary Task Training}\label{sec:hatecheck}

There is an extensive literature on how performance metrics aggregated over the full test set are far from conveying enough information to fully evaluate and compare the strengths and weaknesses of models \cite{ribeiro-etal-2020-beyond}, including for the task of hate speech detection \cite{rottger-etal-2021-hatecheck}. Here, we use the HateCheck test suite in English \cite{rottger-etal-2021-hatecheck} and its recent multilingual version MHC \cite{rottger-etal-2022-multilingual}, which includes our two other target languages, Spanish and Italian. These are test sets covering a wide range of hate speech detection aspects that the authors call \textit{functionalities}, testing detection models with hateful and non-hateful sentences of various styles, vocabulary, syntax and hate speech targets.
All 29 \textit{functionalities} are grouped into 11 classes and 7 protected groups as targets\footnote{We refer the reader to \cite{rottger-etal-2022-multilingual}, pp.45, for an extensive definition of these classes and groups.}, and the various test cases of each functionality lead to a total of 3,901 sentences classified as hateful or not hateful. The protected groups vary across languages in the MHC test set; the authors selected them to better adapt to the cultural context of each language. The target group ``women'' is covered for our three languages, but the target group ``immigrants'' is not covered in Spanish; instead, we match it with the group ``indigenous people''.\footnote{This choice stems from measuring the similarity between Spanish immigrants train set and the test cases of each target group in Spanish Hatecheck using tf-idf representation. Indigenous people (“indígenas” in Spanish) had the highest similarity score with the Twitter immigrants dataset, higher than Hatecheck test cases targeted at black people (“negros”) or Jews (“judíos”), hence our decision to use indigenous people as a proxy. 
} Moreover, to ease the interpretation, we perform the analysis on the aggregated 11 classes of functionalities.

We do not evaluate the performance of our various models on the test suite intrinsically: what we want to measure is the \textit{effect} of zero-shot cross-lingual transfer and auxiliary tasks training on the hate speech functionalities. First, we measure the difference between monolingual and zero-shot cross-lingual training on the various functionalities: what the model ``loses'' by not being trained on the same language as the test set. We rank the functionalities by average difference across the two domains (Table \ref{tab:hatecheck-monoVScross}). 
The largest loss in performance when performing zero-shot transfer is found for functionalities involving slurs: -14.72 of macro-F1 for the immigrants domain  and  -17.22  for the women domain. Indeed, slurs are extremely cultural and language-specific.
\begin{table}[!h]
    \centering
\begin{tabular}{lrr}
\toprule
functionality &  immigrants &  women \\
\midrule
         slur &              -14.72 &         -17.22 \\
       negate &              -10.34 &           0.82 \\
        spell &               -7.56 &           5.78 \\
        derog &               -9.37 &           7.92 \\
       threat &               -2.61 &           1.63 \\
        ident &                5.57 &          -3.22 \\
      counter &               -2.43 &          10.03 \\
          ref &                6.62 &           7.11 \\
    profanity &               -3.75 &          18.33 \\
       phrase &               18.57 &           5.63 \\
\bottomrule
\end{tabular}
\caption{Difference between monolingual and zero-shot cross-lingual performance by functionality when fine-tuning \xlmr on hate speech detection (no auxiliary task), averaged over all language pairs, by domain.}
    \label{tab:hatecheck-monoVScross}
\end{table}
Second, we measure the impact of multilingual auxiliary task training compared to training on hate speech detection only (baseline model), on the various functionalities. For the two domains and for each source-target language pair, we measure the HateCheck functionality score of the baseline model, and jointly on every auxiliary task. For each auxiliary task, we compute the \textit{relative} difference in score with the baseline model; this difference represents the effect of the joint training. However, we focus here on the joint training impact for zero-shot cross-lingual transfer; thus, we  separate the impact of auxiliary task training in a monolingual setting and in a cross-lingual setting. In Table \ref{tab:hatecheck-tasks}, we display the effect of auxiliary task training on zero-shot transfer \textit{on top of} the effect of these tasks on monolingual transfer. To designate the functionalities, we use the same denomination as in the HateCheck test suite. Detection of hate speech involving \textit{slurs}, which suffers the most from zero-shot cross-lingual transfer, is improved by training with NER or UD. Training on UD tasks is especially helpful on cases involving spelling variations (\textit{spell}), contrarily to the two other tasks, and phrasing variations (\textit{phrasing}). Counter-speech detection, an extremely hard task involving \textit{not} classifying counter-speech (e.g. denouncement of hate by quoting it) as hateful, is only helped by NER. Sentiment analysis is globally helpful for many classes, but particularly for sentences involving \textit{negated} positive or hateful statements.

\begin{table}[htbp]
    \centering
\begin{tabular}{lrrr}
\toprule
functionality &  NER &  Sentiment &  UD \\
\midrule
       threat &             -8.23 &                    -2.32 &            26.81 \\
       target &             -3.54 &                     4.70 &            -6.19 \\
        spell &             -3.13 &                    -5.72 &            12.59 \\
         slur &              1.09 &                    -6.30 &            14.42 \\
          ref &             -6.80 &                     2.17 &             7.77 \\
    profanity &             -4.23 &                     2.77 &            -0.44 \\
       phrase &            -14.79 &                     1.17 &             8.64 \\
       negate &              4.19 &                     3.57 &             1.98 \\
        ident &              2.57 &                     1.05 &           -14.42 \\
        derog &             -1.60 &                     2.02 &            18.58 \\
      counter &              2.90 &                   -11.83 &           -15.60 \\
\bottomrule
\end{tabular}
\caption{Relative difference in macro-F1 score by class of functionality, between monolingual and zero-shot cross-lingual training (averaged across all language pairs), averaged across the two domains, for each auxiliary task.}
    \label{tab:hatecheck-tasks}
\end{table}

\section{Discussion}

\paragraph{On the impact of each auxiliary task training,}
we experimented with jointly training hate speech detection and different auxiliary tasks: sentiment analysis, NER and UD tasks. In the immigrants domain, \draftadd{the} NER and sentiment auxiliary tasks led to the best improvement \draftreplace{of}{on} hate speech detection. The cross-lingual transferability of NER was facilitated by the fact that many named entities are the same across languages (e.g. person and organisation names); \draftreplace{besides}{indeed}, many successful unsupervised cross-lingual transfer systems for this task can be found in the literature \cite{rahimi-etal-2019-massively, bari2020zero}. 

Compared with the first two tasks, adding syntactic information had the lowest positive impact on hate speech detection, often decreasing the performance for zero-shot cross-lingual settings. This is in line with results from the literature that agree on the positive effect on sentiment analysis \citep{del2021multi,aroyehun2021evaluation}, but face varying conclusions when it comes to UD tasks. \citet{narang-brew-2020-abusive} showed the positive impact of syntactic features on top of non-contextualized embeddings for hate speech detection; \citet{gambino2020chilab}, among the best systems on the EVALITA2020 hate speech detection task, used POS-tagged text as input for classification. On the contrary, in a monolingual setting, \citet{klemen2020enhancing} showed that morphological features added to LSTM and BERT-based hate speech detection models did not help with comment filtering. Similarly, using sequential auxiliary training of tasks such as POS tagging,  \citet{pruksachatkun-etal-2020-intermediate} showed that the resulting additional low-level skills often led to negative transfer for many downstream tasks. 

In our cross-lingual setting, our goal was to use these tasks as a proxy to fill the mismatch between languages and facilitate the transfer. We hypothesize that when working on tweets, their constrained style -- short sentences, generally with low syntactic complexity -- makes additional syntactic knowledge unhelpful (especially in a more difficult to parse user-generated content context) for a downstream task such as hate speech detection, which benefits more from semantic information.

\paragraph{Regarding the non-usage of POS taggers that could have been optimized for our User-Generated Content-based datasets,} we investigated this possibility and conducted preliminary experiments for English – using the Tweebank \cite{jiang-etal-2022-annotating} as data source–, that showed that using a tagger trained on it did not bring much in terms of performance compared to “classic” UD POS taggers. Part of the reasons might come from the fact that our pre-processing step removes hashtags and normalized other Twitter’s idiosyncrasies and hence make the data somewhat simpler to tag. Another reason to not investigate this further lies in the lack of availability of a UGC treebank for Spanish, breaking thus the symmetry of our experimental protocol. Last but not least, another reason we hypothesized for this lack of much improvement we noticed comes from the fact that the multilingual language model we used (XMLR and XMLR-T) were already providing strong results on UGC. This was corroborated by \citet{riabi-etal-2021-character}, who experimentally verified the robustness of language models when facing noisy UGC. Moreover \citet{itzhak2021models} showed that subword-based language models were able to capture a significant amount of character-level alteration typical of UGC \cite{DBLP:journals/corr/abs-2011-02063}, explaining their surprising level of robustness when facing noisy content.
  However, we agree that better handling UGC content would be an interesting step, if not the next step, especially if we can demonstrate that many idiosyncrasies align across languages in our target domains and hence are alleviated by the use of optimized tagging and parsing, eventually multilingual, models. This, in our minds, warrants another full-scale study with a thorough error analysis of cross-lingual syntactic transfer in noisy scenarios. We leave this for future work.

\paragraph{Cross-lingual zero-shot transfer on a domain with a gap between languages.}

In Section \ref{sec:bottleneck}, we observed that using larger pre-trained multilingual language models, and if possible, multilingual models trained on corpora from the same source as the downstream task, improves cross-lingual zero-shot transfer. This adaptation has a significant and consistent positive impact. This is in line with the findings of \citet{bose-etal-2021-unsupervised}, who demonstrated the superiority of MLM over other tasks in a cross-corpora transfer setting. Similarly, \citet{van-der-goot-etal-2021-masked} jointly trained auxiliary tasks with a downstream task (in their case, spoken language understanding) in a cross-lingual setting to find that MLM fine-tuning consistently improves the downstream task. 

Beyond the obvious improvement due to the MLM training on more adapted data, we would have expected \xlmt to increase the impact of auxiliary tasks fine-tuning; a more adapted language model helping to bridge the gap between hate speech in the source and target languages. Here, the Twitter data used for the \xlmt training may not be optimal for the observed linguistic specificities and cultural gap. It was trained on tweets published between 05/2018 and 03/2020, while the hate speech corpora range from 2017 to 2018, depending on the language; moreover, some events were specifically targeted when  scraping Twitter for hate speech detection (e.g., Gamergate victims for the Italian datasets on hate speech towards women \cite{fersini2018overview}). 
Furthermore, contrarily to Wikipedia where corpora are highly similar from one high-resource language to another in term of domains, Twitter data can significantly differ between languages due to cultural differences and events in the respective countries.
Overall, when we used \xlmt, the model is only adapted to the form and style of Twitter data (small sentences, with mentions and urls\dots). The tweets' content, topic, and vocabulary might differ a lot between the hate speech corpora, the {\xlmt} training data, and the sentiment analysis corpora. We can only hypothesize on these variations. However, they should be quantified to understand better the impact of fine-tuning on these data and to distinguish between corpus variations and the actual cultural and linguistic gap.

Discussions on computational costs and ethical considerations for this work can be found in Appendix \ref{app:ethical-consideration}.




\section{Conclusion}

In this work, we highlighted situations where zero-shot cross-lingual transfer of hate speech models fails because of the linguistic and cultural gap. We quantified the effect of the choice of multilingual language model and of auxiliary task training on these ``failed'' cases, showing the positive effect of NER and sentiment analysis multilingual training, but their limited improvement in the domain of hate speech against women. We performed a preliminary analysis on the effect of auxiliary tasks by \textit{hate speech functionality} using the HateCheck test suite, hinting at which kind of hate speech benefits from transferring knowledge in both the source and the target languages for the three auxiliary tasks.
 Finally, we discussed limitations related to training data for language model pre-training, auxiliary tasks, and hate speech detection. All of our datasets with their new splits and models are freely available.\footnote{\url{https://github.com/ArijRB/Multilingual-Auxiliary-Tasks-Training-Bridging-the-Gap-between-Languages-for-Zero-Shot-Transfer-of-/}}, hoping that the sound experimental framework we designed will help strengthen future studies on cross-lingual hate-speech detection.
\section*{Acknowledgments}
We warmly thank the reviewers for their very valuable feedback.
This work  received
funding from the European Union’s Horizon 2020
research and innovation programme under grant
agreement No. 101021607. The last author acknowledges the support of the French Research Agency via the ANR ParSiTi project (ANR16-CE33-0021).
The authors are grateful to the OPAL infrastructure from Université Côte d'Azur for providing resources and support.

\newpage
\section{Ethical considerations}\label{app:ethical-consideration}

This paper is part of a line of work aiming to tackle hate speech detection when we have no training data in the target language, fight the spread of offensive and hateful speech online, and have a positive global impact on the world. Its goal is to understand if hate speech is transferable from one language to another; as such, it has been approved by our institutional review board (IRB), and follows the national and European General Data Protection Regulation (GDPR).

We did not collect any data from online social media for this work. We only used publicly available datasets -- exclusively diffused for shared tasks that were tackled by a large number of participants (see Table \ref{tab:shared_hate} in Appendix \ref{sec:appendix-datasets}). These datasets do not include any metadata, only the tweet's text associated with the hate speech label. Thus, linking the annotated data to individual social media users is not straightforward.

All our experiments were executed on clusters whose energy mix is made of nuclear (65--75\%), 20\% renewable, and the remaining with gas (or more rarely coal when imported from abroad). More details on computational costs can be found in Table \ref{tab:duration_experiments}.

Finally, the presence of bias in the pre-trained language models we use, due to the bias in the data they were trained on, may have an impact on hate speech detection, particularly on the topic of hate speech towards women. As a result, this area of research is currently under heavy scrutiny by the community.

 \paragraph{Computational Costs.} 
We conduct our experiments on RTX8000 GPUs. We test two models (\xlmr and \xlmt) on 7 different auxiliary tasks combinations, with 5 seeds each. Details on the average GPU time for the basic task combinations (jointly training hate speech with one task) are in Table \ref{tab:duration_experiments}.
 
 \begin{table}[!h]
    \centering
    \footnotesize
    \begin{tabular}{lr}
    \toprule
   Task  & Duration\\
    \midrule
    Hate only&   0:14\\
    Sentiment+Hate&   0:21\\
    UD+Hate&   1:57\\
    NER+Hate&   2:18\\
    \bottomrule
    \end{tabular}
    \caption{Training time (in seconds) for one seed per model.}
    \label{tab:duration_experiments}
\end{table}

\bibliography{anthology,custom}
\bibliographystyle{acl_natbib}


\appendix

\section{Datasets overview}\label{sec:appendix-datasets}

\subsection{Hate speech datasets overview}

\begin{table*}[!ht]
    \centering
    \footnotesize
    \begin{tabular}{llrrr}
    \toprule
    Shared task & Link &   \\
    \midrule
 Hateval  & \url{https://github.com/msang/hateval} \\
   EVALITA AMi 2018 & \url{https://github.com/MIND-Lab/ami2018} \\
  HaSpeeDe 2018  & \url{https://github.com/msang/haspeede/tree/master/2018} \\

    \bottomrule
    \end{tabular}
    \centering
    \caption {Shared tasks used for the Hate speech corpora.}
    \label{tab:shared_hate}
\end{table*}

\begin{table}
    \centering
    \footnotesize
    \begin{tabular}{lrrrc}
    \toprule
    Domain-language & train & dev & test & blind \\
    \midrule
    immigrants-it & 2000 & 500 & 1000 & . \\
    immigrants-en & 4500 & 500 & 1499 & . \\
    immigrants-es & 1618 & 173 & 800 & . \\
    women-it & 2500 & 500 & 1000 & . \\
    women-en & 4500 & 500 & 1472 & . \\
    women-es & 2882 & 327 & 799 & . \\
    \midrule    
    Comparable size & 1618 & 173 & 800 & 1000 \\
    \bottomrule
\end{tabular}
\caption{Hate speech detection datasets: Size of full datasets (number of sentences) and new split with comparable data size. Only the immigrants-es dataset has no blind set.}
\label{tab:hatespeech_data_size}
\end{table}

\begin{table}[!ht]
    \centering
    \footnotesize
    \begin{tabular}{lrr}
    \toprule
   Language & immigrants & women    \\
    \midrule
    en & 41.28 &  42.76 \\
    es & 42 &  40.23 \\
    it & 31.33 &  45.42 \\
    \bottomrule
    \end{tabular}
    \caption{Percentage of hateful examples in the train set for the comparable setting.}
    \label{tab:hatespeech_percentage}
\end{table}

\begin{table}[!ht]
\centering
\footnotesize
\begin{tabular}{lrrrrrr}
\toprule
{} & \multicolumn{3}{c}{immigrants} & \multicolumn{3}{c}{women} \\
\cline{2-7}
 &en &es &it &en &es &it \\\midrule
 \multicolumn{7}{c}{Nb of tokens per tweet}  \\\midrule
avg     &          27.3 &          18.9 &          17.2 &     18.3 &     22.8 &     17.9 \\
median   &          26.0 &          17.0 &          17.0 &     18.0 &     20.0 &     14.0 \\
max    &            90 &            57 &            29 &       57 &       59 &       54 \\
min   &             2 &             1 &             2 &        2 &        2 &        2 \\\midrule
 \multicolumn{7}{c}{Nb of hashtags (avg per tweet, total unique nb)}  \\\midrule
avg  &           2.0 &           0.2 &           0.6 &      0.2 &      0.2 &      0.2 \\
unique  &          1162 &           214 &           491 &      211 &      292 &      228 \\
\midrule
\multicolumn{7}{c}{Train/test OOV Ratio }  \\\midrule
                &           0.4 &           0.6 &           0.5 &      0.5 &      0.5 &      0.5 \\
\bottomrule
\end{tabular}

\caption{Descriptive statistics on hate speech detection training datasets.}
\label{tab:hatespeech_data_stat}
\end{table}

\subsection{Auxiliary tasks datasets overview}

\begin{table*}[!ht]
    \centering
    \footnotesize
    \begin{tabular}{llll}
    \toprule
   Language & Shared task & Reference  & Scraping period  \\
    \midrule
    English & SemEval 2017 & \citet{rosenthal2017semeval} &  01/2012--12/2015 \\
     Italian & Intertass 2017  & \citet{diaz2018democratization} &  07/2016--01/2017 \\
    Spanish & Sentipolc 2016 & \citet{barbieri2016overview} &   2013--2016 \\
    \bottomrule
    \end{tabular}
    \caption{Data overview for the sentiment analysis task. All datasets contain text scraped from Twitter. They have been unified to a common train / dev / test split size: 1\,839 / 324 / 870.}
    \label{tab:complementary-datasets-sentiment}
\end{table*}

\begin{table*}[!ht]
    \centering
    \footnotesize
    \begin{tabular}{cccc}
    \toprule
    Dataset & Language &  train/dev/test size & Period\\
    \midrule
  Tweebank & English & 1\,639 / 710 / 1\,201 & 02/2016 -- 07/2016\\
  PoSTWITA & Italian & 5\,368 / 671 / 674 & 07/2009 -- 02/2013 \\
    \bottomrule
    \end{tabular}
    \caption{Twitter UD data overview.}
    \label{tab:UD-twitter-data}
\end{table*}

\begin{table}[!ht]
    \centering
    \footnotesize
    \begin{tabular}{llrrr}
    \toprule
    Dataset & Language &  train & dev & test  \\
    \midrule
  EWT\footnote{\url{https://github.com/UniversalDependencies/UD_English-EWT/}} & English & 12\,543 & 2\,001 & 2\,077 \\
   GSD\footnote{\url{https://github.com/ryanmcd/uni-dep-tb}, \url{https://github.com/UniversalDependencies/UD_Spanish-GSD/}} & Spanish &  14\,187 & 1\,400 & 426\\
   ISDT\footnote{\url{https://github.com/UniversalDependencies/UD_Italian-ISDT/}} & Italian & 13\,121 & 564 & 482\\
   \midrule
   \multicolumn{2}{l}{Comparable size} & 12543 & 564 & 426\\
    \bottomrule
    \end{tabular}
    \caption{Universal Dependencies (UD) datasets and size of their respective splits.}
    \label{tab:complementary-datasets-ud}
\end{table}

\begin{table}[!ht]
    \centering
    \footnotesize
    \begin{tabular}{ccc}
    \toprule
     & Train &  Dev \\
    \midrule
   \# tweets & 2\,349 & 1\,000  \\
   \# tokens & 46\,469 & 16\,261  \\
   \# entity tokens & 2\,462 &  1\,128\\
    \bottomrule
    \end{tabular}
    \caption{Statistics of the WNUT 2016 NER shared task dataset.}
    \label{tab:NER-twitter-data}
\end{table}

\paragraph{Treebanks additional pre-processing}

As the \machamp framework does not support the Connl UD format, treebanks must be converted back to the connl06 format, which most notably involved the removal of all contracted tokens, potentially leading to tokenization mismatches between our data sources. However, a rapid analysis showed that it has a very limited impact because of their low frequency and the generalization of sub-word tokenization.

\clearpage
\newpage

\section{Complementary results}\label{app:results}

\begin{table}
\footnotesize
\begin{subtable}{1\linewidth}
\sisetup{table-format=-1.2}   
\centering

    \begin{tabular}{@{}p{0.5cm}p{0.4cm}cccccc@{}}
    \toprule

\multirow{2}{*}{\parbox{0.6cm}{\centering Aux task}} & \multirow{2}{*}{\parbox{0.5cm}{\centering  Src lang}} &\multicolumn{3}{c}{immigrants} & \multicolumn{3}{c}{women} \\ \cmidrule(lr){3-5}
\cmidrule(l){6-8}
& &en &es &it &en &es &it \\ \midrule
\multirow{3}{*}{None}& en  & 76.8\hphantom{$^\dagger$} & 48.5\hphantom{$^\dagger$} & 73.5\hphantom{$^\dagger$} & 78.6\hphantom{$^\dagger$} & 61.5\hphantom{$^\dagger$} & 60.6\hphantom{$^\dagger$} \\ 
 & es  & 65.9\hphantom{$^\dagger$} & 84.2\hphantom{$^\dagger$} & 60.7\hphantom{$^\dagger$} & 72.5\hphantom{$^\dagger$} & 80.3\hphantom{$^\dagger$} & 51.9\hphantom{$^\dagger$} \\ 
 & it  & 71.5\hphantom{$^\dagger$} & 56.8\hphantom{$^\dagger$} & 78.4\hphantom{$^\dagger$} & 63.4\hphantom{$^\dagger$} & 58.2\hphantom{$^\dagger$} & 90.3\hphantom{$^\dagger$} \\ 

\midrule
\multirow{3}{*}{sent} & en  & \textcolor{red!60!black}{-0.4\hphantom{$^\dagger$}} & \textcolor{green!60!black}{4.2$^\dagger$} & \textcolor{red!60!black}{-1.9\hphantom{$^\dagger$}} & \textcolor{green!60!black}{0.5\hphantom{$^\dagger$}} & \textcolor{green!60!black}{2.2\hphantom{$^\dagger$}} & \textcolor{red!60!black}{-0.2\hphantom{$^\dagger$}} \\ 
 & es  & \textcolor{green!60!black}{1.3\hphantom{$^\dagger$}} & \textcolor{green!60!black}{0.5\hphantom{$^\dagger$}} & \textcolor{green!60!black}{6.2\hphantom{$^\dagger$}} & \textcolor{red!60!black}{-2.6$^\dagger$} & \textcolor{green!60!black}{0.7\hphantom{$^\dagger$}} & \textcolor{red!60!black}{-9.6$^\ddagger$} \\ 
 & it  & \textcolor{green!60!black}{0.8\hphantom{$^\dagger$}} & \textcolor{red!60!black}{-1.8\hphantom{$^\dagger$}} & \textcolor{red!60!black}{-0.3\hphantom{$^\dagger$}} & \textcolor{red!60!black}{-5.1$^\dagger$} & \textcolor{green!60!black}{3.4\hphantom{$^\dagger$}} & \textcolor{red!60!black}{-0.3\hphantom{$^\dagger$}} \\ 
\midrule 
\multirow{3}{*}{NER}   & en  & \textcolor{green!60!black}{0.1\hphantom{$^\dagger$}} & \textcolor{green!60!black}{5.9$^\ddagger$} & \textcolor{red!60!black}{-4.7$^\ddagger$} & \textcolor{red!60!black}{-0.1\hphantom{$^\dagger$}} & \textcolor{green!60!black}{0.9\hphantom{$^\dagger$}} & \textcolor{green!60!black}{1.6\hphantom{$^\dagger$}} \\ 
 & es  & \textcolor{red!60!black}{-2.2\hphantom{$^\dagger$}} & \textcolor{green!60!black}{0.6\hphantom{$^\dagger$}} & \textcolor{green!60!black}{1.4\hphantom{$^\dagger$}} & \textcolor{red!60!black}{-5.9$^\ddagger$} & \textcolor{green!60!black}{1.5$^\dagger$} & \textcolor{red!60!black}{-6.0$^\ddagger$} \\ 
 & it  & \textcolor{green!60!black}{1.0\hphantom{$^\dagger$}} & \textcolor{green!60!black}{0.7\hphantom{$^\dagger$}} & \textcolor{green!60!black}{0.0\hphantom{$^\dagger$}} & \textcolor{red!60!black}{-2.7\hphantom{$^\dagger$}} & \textcolor{green!60!black}{2.2\hphantom{$^\dagger$}} & \textcolor{green!60!black}{0.5\hphantom{$^\dagger$}} \\

\midrule 
\multirow{3}{*}{UD} & en  & \textcolor{red!60!black}{-0.4\hphantom{$^\dagger$}} & \textcolor{green!60!black}{2.9\hphantom{$^\dagger$}} & \textcolor{red!60!black}{-3.9\hphantom{$^\dagger$}} & \textcolor{red!60!black}{-0.1\hphantom{$^\dagger$}} & \textcolor{red!60!black}{-1.7$^\dagger$} & \textcolor{red!60!black}{-10.1$^\ddagger$} \\ 
 & es  & \textcolor{red!60!black}{-11.1$^\ddagger$} & \textcolor{red!60!black}{-0.7\hphantom{$^\dagger$}} & \textcolor{red!60!black}{-3.7\hphantom{$^\dagger$}} & \textcolor{red!60!black}{-2.4$^\ddagger$} & \textcolor{green!60!black}{0.4\hphantom{$^\dagger$}} & \textcolor{red!60!black}{-12.9$^\ddagger$} \\ 
 & it  & \textcolor{red!60!black}{-4.1$^\ddagger$} & \textcolor{green!60!black}{1.6\hphantom{$^\dagger$}} & \textcolor{green!60!black}{0.1\hphantom{$^\dagger$}} & \textcolor{red!60!black}{-8.7$^\ddagger$} & \textcolor{red!60!black}{-2.1\hphantom{$^\dagger$}} & \textcolor{green!60!black}{0.7$^\dagger$} \\

\bottomrule

    \end{tabular}%

   \caption{Detailed view.}\label{tab:xlmt-detail}
\end{subtable}

\bigskip
\begin{subtable}{1\linewidth}
\sisetup{table-format=4.0} 
\centering

    \begin{tabular}{lrrrr}
    \toprule 
    \multirow{2}{*}{\makecell{Auxiliary\\ Task}} &  \multicolumn{2}{c}{immigrants} &  \multicolumn{2}{c}{women} \\ 
    \cmidrule(lr){2-3}
    \cmidrule(lr){4-5}
     & mono & cross & mono & cross \\  
      \midrule 
 None  & 79.8 & 62.8 & 83.1 & 61.3 \\ 
Sent  & \textcolor{red!60!black}{-0.1} & \textcolor{green!60!black}{1.5} & \textcolor{green!60!black}{0.3} & \textcolor{red!60!black}{-2.0} \\ 
NER  & \textcolor{green!60!black}{0.3} & \textcolor{green!60!black}{0.4} & \textcolor{green!60!black}{0.6} & \textcolor{red!60!black}{-1.7} \\ 
UD  & \textcolor{red!60!black}{-0.3} & \textcolor{red!60!black}{-3.0} & \textcolor{green!60!black}{0.3} & \textcolor{red!60!black}{-6.3} \\ 
Sent + NER  & \textcolor{red!60!black}{-0.2} & \textcolor{green!60!black}{1.3} & \textcolor{green!60!black}{0.6} & \textcolor{red!60!black}{-2.5} \\ 
\bottomrule 

    \end{tabular}%
\caption{Aggregated view.}
\label{tab:xlmt-summary}
\end{subtable}

    \caption[]{Effect (delta with hate speech detection baseline, averaged over 5 runs) of fine-tuning \xlmt on the three auxiliary tasks, on hate speech detection macro-F1 scores (\%). \small{\textcolor{green!60!black}{Green} values indicate an increase in score, \textcolor{red!60!black}{red} values a decrease. \textit{Sent} stands for Sentiment and \textit{Aux} for auxiliary. }}
    \label{tab:xlmt-all}
\end{table}

\begin{table}
\centering
    \footnotesize
   \begin{tabular}{llrrr}
    \toprule

{\parbox{0.5cm}{\centering Aux. task}} & {\parbox{0.5cm}{\centering  Src lang}} &en &es  &it\\ \midrule

 \multirow{3}{*}{None }& en  & \textit{75.3} & \textcolor{orange!80!black}{51.9} & 70.1 \\ 
 & es  & \textcolor{orange!80!black}{62.0} & \textit{83.4} & 65.4 \\ 
 & it  & 69.2 & \textcolor{orange!80!black}{51.3} & \textit{78.6} \\ 
\midrule 
 \multirow{3}{*}{MLM}& en & \textcolor{green!60!black}{1.1\hphantom{$^\dagger$}} & \textcolor{red!60!black}{-2.9\hphantom{$^\dagger$}} & \textcolor{red!60!black}{-1.4\hphantom{$^\dagger$}} \\ 
 & es  & \textcolor{green!60!black}{2.6\hphantom{$^\dagger$}} & \textcolor{red!60!black}{-2.9$^\ddagger$} & \textcolor{green!60!black}{0.3\hphantom{$^\dagger$}} \\ 
& it  & \textcolor{red!60!black}{-1.6\hphantom{$^\dagger$}} & \textcolor{red!60!black}{-1.0\hphantom{$^\dagger$}} & \textcolor{red!60!black}{-0.1\hphantom{$^\dagger$}}  \\ 
\midrule 
\multirow{3}{*}{NER} & en  & \textcolor{green!60!black}{1.4$^\dagger$} & \textcolor{green!60!black}{1.0\hphantom{$^\dagger$}} & \textcolor{red!60!black}{-1.9\hphantom{$^\dagger$}}   \\ 
 & es  & \textcolor{green!60!black}{3.1\hphantom{$^\dagger$}} & \textcolor{green!60!black}{0.4\hphantom{$^\dagger$}} & \textcolor{red!60!black}{-1.1\hphantom{$^\dagger$}}  \\ 
 & it  & \textcolor{green!60!black}{3.3$^\ddagger$} & \textcolor{green!60!black}{4.5$^\ddagger$} & \textcolor{red!60!black}{-1.4$^\dagger$}  \\

\bottomrule
    \end{tabular}
    \caption{Effect (delta with XLM-R baseline) of MLM fine-tuning on sentences from NER datasets compared fine-tuning on NER as auxiliary tasks, on hate speech detection macro-F1 scores (\%) for immigrants domain. \small{\textcolor{green!60!black}{Green} values indicate an increase in score, \textcolor{red!60!black}{red} values a decrease. } }
    \label{tab:xlmr-mlm-ner}
\end{table}

\begin{table*}[!ht]
    \centering
    \footnotesize
    \begin{tabular}{llcccccc}
    \toprule
\multirow{2}{*}{\parbox{1cm}{\centering Auxiliary task}} & \multirow{2}{*}{\parbox{1cm}{\centering  Source lang}} & \multicolumn{3}{c}{immigrants} & \multicolumn{3}{c}{women} \\
\cline{3-8}
 & &en &es &it &en &es &it \\\midrule

\multirow{3}{*}{None }& en  & \textit{75.3} & \textcolor{orange!80!black}{51.9} & 70.1 & \textit{76.6} & \textcolor{orange!80!black}{51.6} & \textcolor{orange!80!black}{49.9} \\ 
 & es  & \textcolor{orange!80!black}{62.0} & \textit{83.4} & 65.4 & 63.4 & \textit{77.8} & \textcolor{orange!80!black}{46.9} \\ 
 & it  & 69.2 & \textcolor{orange!80!black}{51.3} & \textit{78.6} & \textcolor{orange!80!black}{60.3} & \textcolor{orange!80!black}{57.3} & \textit{89.0} \\ 
 \midrule 
\multirow{3}{*}{UD} & en  & \textcolor{green!60!black}{1.7$^\dagger$} & \textcolor{red!60!black}{-2.4\hphantom{$^\dagger$}} & \textcolor{red!60!black}{-1.2\hphantom{$^\dagger$}} & \textcolor{green!60!black}{0.7\hphantom{$^\dagger$}} & \textcolor{red!60!black}{-0.4\hphantom{$^\dagger$}} & \textcolor{red!60!black}{-10.6$^\dagger$} \\ 
 & es  & \textcolor{red!60!black}{-3.6\hphantom{$^\dagger$}} & \textcolor{red!60!black}{-1.1\hphantom{$^\dagger$}} & \textcolor{red!60!black}{-6.5$^\dagger$} & \textcolor{red!60!black}{-4.9\hphantom{$^\dagger$}} & \textcolor{red!60!black}{-0.4\hphantom{$^\dagger$}} & \textcolor{red!60!black}{-10.9$^\ddagger$} \\ 
 & it  & \textcolor{red!60!black}{-14.4$^\ddagger$} & \textcolor{green!60!black}{5.0$^\ddagger$} & \textcolor{red!60!black}{-1.6$^\dagger$} & \textcolor{red!60!black}{-14.7$^\ddagger$} & \textcolor{red!60!black}{-5.6\hphantom{$^\dagger$}} & \textcolor{red!60!black}{-0.3\hphantom{$^\dagger$}} \\ 
 \midrule 
\multirow{3}{*}{UPOS}  & en  & \textcolor{red!60!black}{-0.6\hphantom{$^\dagger$}} & \textcolor{red!60!black}{-3.1\hphantom{$^\dagger$}} & \textcolor{red!60!black}{-1.4\hphantom{$^\dagger$}} & \textcolor{green!60!black}{0.9\hphantom{$^\dagger$}} & \textcolor{red!60!black}{-5.2\hphantom{$^\dagger$}} & \textcolor{red!60!black}{-1.2\hphantom{$^\dagger$}} \\ 
 & es  & \textcolor{red!60!black}{-4.0\hphantom{$^\dagger$}} & \textcolor{red!60!black}{-1.2\hphantom{$^\dagger$}} & \textcolor{red!60!black}{-3.9$^\dagger$} & \textcolor{red!60!black}{-0.9\hphantom{$^\dagger$}} & \textcolor{green!60!black}{1.9$^\ddagger$} & \textcolor{red!60!black}{-7.3$^\dagger$} \\ 
 & it  & \textcolor{red!60!black}{-4.7$^\dagger$} & \textcolor{green!60!black}{5.0$^\ddagger$} & \textcolor{red!60!black}{-1.0\hphantom{$^\dagger$}} & \textcolor{red!60!black}{-1.2\hphantom{$^\dagger$}} & \textcolor{red!60!black}{-3.4\hphantom{$^\dagger$}} & \textcolor{red!60!black}{-1.7\hphantom{$^\dagger$}} \\

\bottomrule
    \end{tabular}
    \caption{\textbf{Ablation study}: Hate speech detection macro-F1 scores (\%) of \xlmr fine-tuned on the UPOS task jointly with the hate speech detection task. We compare each macro-F1 score with the baseline score (without auxiliary task). \textcolor{green!60!black}{Green} values indicate an increase in score, \textcolor{red!60!black}{red} values a decrease. The subscript indicates whether the macro-F1 of the model trained with the auxiliary tasks is significantly higher or lower compared to the model without auxiliary task. The comparison is made using a one-sided $t$-test over the list of scores of the five runs of each model. A dagger ($\dagger$) as exponent indicates that the $p$-value is smaller than 0.05, while a double-dagger ($\ddagger$) indicates a $p$-value smaller than 0.01.}
    \label{tab:xlmr-ud_ablation}
\end{table*}
\clearpage
\newpage


\end{document}